%% file: def_aistats_arxiv.tex
\documentclass[twoside]{article}
\usepackage[compact]{titlesec} 
\usepackage{enumitem}
\usepackage{times}

\usepackage[accepted]{aistats2015}
\input{preamble}

\usepackage[numbers]{natbib}
\usepackage{algorithm}
\usepackage{algorithmic}
\usepackage{subcaption}
\usepackage{framed}
\usepackage{bm}
\usepackage{url}

\usepackage{array}
\newcolumntype{"}{@{\hskip\tabcolsep\vrule width 1.5pt\hskip\tabcolsep}}
\makeatother

\begin{document}

%

%

\twocolumn[

\aistatstitle{Deep Exponential Families}

\aistatsauthor{ Rajesh Ranganath \And Linpeng Tang \And Laurent Charlin \And David M. Blei}

\aistatsaddress{ Princeton University \And Princeton University \And Columbia University \And Columbia University \AND
{\tt $\{$rajeshr,linpengt$\}$@cs.princeton.edu} \\
{\tt $\{$lcharlin,blei$\}$@cs.columbia.edu}  }

]

\begin{abstract}
  We describe \textit{deep exponential families} (DEFs), a class of latent
  variable models that are inspired by the hidden structures used in
  deep neural networks. 
  DEFs capture a hierarchy of dependencies between latent variables, and
  are easily generalized to many settings through exponential families.
  We perform inference 
  using recent ``black box" variational inference techniques. We then evaluate 
  various DEFs on text and combine multiple DEFs into a
  model for pairwise recommendation
  data. In an extensive study, we show that going beyond one layer improves predictions for DEFs.
  We demonstrate that DEFs find interesting exploratory structure in
  large data sets, and give better predictive performance than
  state-of-the-art models.
\end{abstract}

\section{Introduction}

In this paper we develop deep exponential families (DEFs), a flexible
family of probability distributions that reflect the intuitions behind
deep unsupervised feature learning. In a DEF, observations arise from
a cascade of layers of latent variables. Each layer's variables are
drawn from an exponential family that is governed by the inner product
of the previous layer's variables and a set of weights. 

As in deep unsupervised feature learning, a DEF represents
hidden patterns, from coarse to fine grained, that compose with each other
to form the observations.
DEFs also enjoy the advantages of probabilistic modeling. Through
their connection to exponential families~\cite{Brown:1986}, they
support many kinds of data. 
Overall DEFs combine the powerful representations of deep networks with
the flexibility of graphical models.

Consider the problem of modeling documents. We can
represent a document as a vector of term counts modeled with 
Poisson random variables~\citep{Canny:2004}. In one type of DEF,
the rate of each term's Poisson count is
an inner product of a layer of latent variables (one level up from the
terms) and a set of weights that are shared across documents. Loosely,
we can think of the latent layer above the observations as per-document
``topic'' activations, each of
which ignites a set of related terms via their inner product with
the weights. These latent topic are, in turn, modeled in a similar
way, conditioned on a layer above of ``super topics.'' Just as the
topics group related terms, the super topics group related topics,
again via the inner product.

\begin{figure*}
	\centering
	    \includegraphics[width=\textwidth]{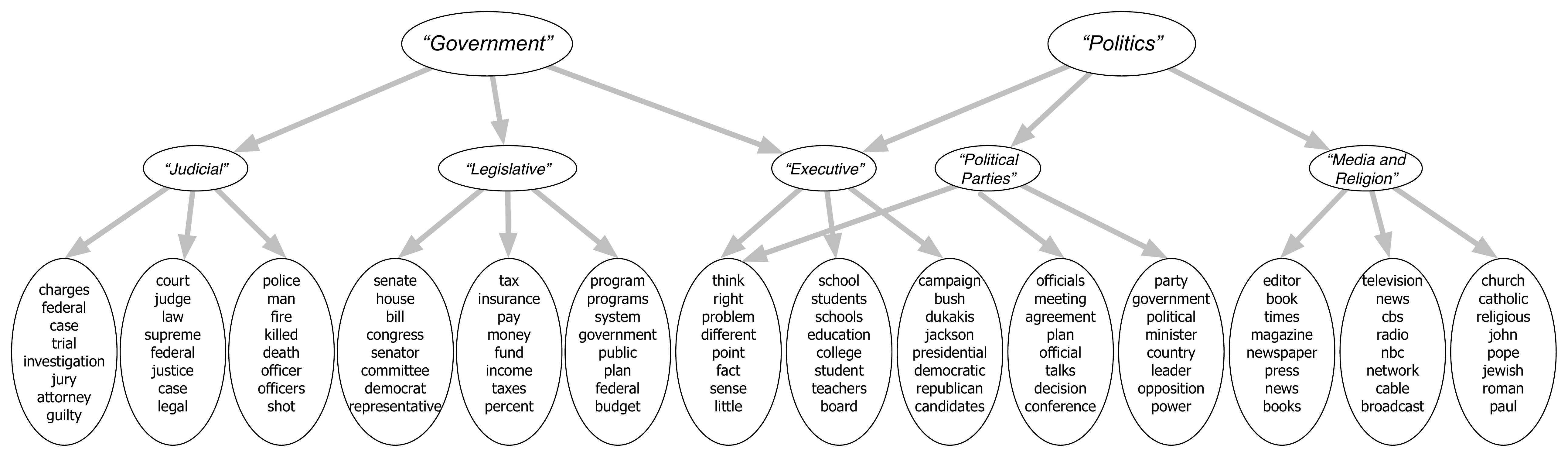}
	\caption{A fraction of the three layer topic hierarchy on 166K \emph{The New York Times} articles. 
				The top words are shown for each topic.
	   			The arrows represent hierarchical groupings.}
	\label{fig:nyt_topics}
\end{figure*}

\myfig{nyt_topics} illustrates an example of a three level DEF uncovered 
from a large set of articles in The New York Times.  (This style of model, though with
different details, has been previously studied in the topic modeling
literature \citep{Li:2006}.)  Conditional on the word counts of the articles, the
DEF defines a posterior distribution of the per-document cascades of
latent variables and the layers of weights.  Here we have visualized
two third-layer topics which correspond to the \emph{concepts} of
``Government'' and ``Politics''. We focus on ``Government" and notice
that the model has discovered, through its second-layer
\emph{super-topics}, the three branches of government: judiciary (
left), legislative (center) and executive (right).

This is just one example.  In a DEF, the latent
variables can be from any exponential family: Bernoulli latent
variables recover the classical sigmoid belief
network~\cite{Neal:1990}; Gamma latent variables give something akin
to deep version of nonnegative matrix factorization~\cite{Lee:1999}; Gaussian latent
variables lead to the types of models that have recently been explored
in the context of computer vision~\cite{Rezende:2014}.  DEFs fall 
into the broad class of stochastic feed forward networks defined
by~\citet{Neal:1990}. These networks differ from the undirected 
deep probabilistic models \citep{Salakhutdinov:2009b,Srivastava:2013} in that they
allow for explaining away, where latent variables compete
to explain the observations. 

In addition to varying types of latent variables, we can further change the
prior on the weights and the observation model. Observations can be real valued, such as those
from music and images, binary, such as those
in the sigmoid belief network, or multinomial, such as
when modeling text. In the language of
neural networks, the prior on the weights amounts to choosing a type
of regularization; the observation model amounts to choosing a type of
loss. 

Finally, we can embed the DEF in a more complex model, building "deep"
versions of traditional models from the statistics and machine
learning research literature.
As examples, the DEF can be made part of a
multi-level model of grouped data~\cite{Gelman:2007}, time-series
model of sequential data~\cite{Blei:2006d}, or a factorization model of pairwise
data~\cite{Salakhutdinov:2008}. 
As a concrete example, we will develop and study the \textit{double
DEF}.  The double DEF models a matrix of pairwise observations, such
as users rating items.  It uses two DEFs, one for the latent
representation of a user and one for the representation of an item.  The observation of each
user/item interaction combines the lowest layer of their individual
DEF representations.

In the rest of this paper, we will define, develop, and study deep
exponential families.  We will explain some of their properties and
situate them in the larger contexts of probabilistic models and deep
neural networks.  We will then develop generic variational inference
algorithms for using DEFs. We will s
how how to use them to solve real-world problems with large data
sets, and we will extensively study many DEFs on the problems of
document modeling and collaborative filtering.  We show that
DEF-variants of existing "shallow" models give more interesting
exploratory structure and better predictive performance.  More
generally, DEFs are a flexible class of models which, along with our
algorithms for computing with them, let us easily explore a rich
landscape of solutions for modern data analysis problems.

%

\section{Deep exponential families}
In this section we review exponential families and present deep 
exponential families.

\paragraph{Exponential families.}
Exponential families \citep{Brown:1986} are an important
class of distributions with
convenient mathematical properties. They take the following
form. 
\begin{align*}
p(x) = h(x) \exp (\eta^\top T(x) - a(\eta)),
\end{align*}
where $h$ is the base measure, $\eta$ are the natural
parameters, $T$ are the sufficient statistics, and $a$
is the log-normalizer. The expectation of the sufficient
statistics of an exponential family is the
gradient of the log-normalizer $\E[T(x)] = \nabla_\eta a(\eta)$. Exponential families 
are completely specified by their sufficient statistics
and base measure; different choices of $h$ and $T$
lead to different distributions. For example, in the normal 
distribution the base measure is $h=\sqrt{(2\pi)}$  and the sufficient 
statistics are $T(x)=[x, x^2]$; and for the Beta distribution, a distribution 
with support over $(0,1)$,  the base measure is $h = 1$ and sufficient 
statistics are and $T(x) = [\log x, \log 1 - x]$.

\paragraph{Deep exponential families.}
To construct deep exponential families, 
we will chain exponential families together in a hierarchy,
where the draw from one layer controls the natural parameters
of the next. 


For each data point $x_n$, the model has $L$ layers of hidden
variables $\{ \defbold{z}_{n, 1}, ...,   \defbold{z}_{n, L} \}$, 
where each $\defbold{z}_{n,\ell} = \{z_{n, \ell, 1}, ..., z_{n, \ell, K_\ell} \}$. 
We assume that
$z_{n, \ell, k}$ is a scalar, but the model generalizes beyond this.
Shared across data, the model has $L-1$ layers of weights $\{\defbold{W}_1, ... \defbold{W}_{L-1} \}$,
where each $\defbold{W}_{\ell}$ is a collection of $K_{\ell}$ vectors, each one
with dimension $K_{\ell+1}$: $\defbold{W}_{\ell} = \{\defbold{w}_{\ell, 1}, ... \defbold{w}_{\ell, K_\ell} \}$.
We assume the weights have a prior 
distribution $p(\defbold{W}_{\ell})$.

For simplicity, we omit the data index $n$ and describe the
distribution of a single data point $x$. First, the top layer of 
latent variables are drawn given a hyperparameter $\eta$
\begin{align*}
  p(z_{L,k}) =
  \expfam_L(z_{L, k}, \eta),
\end{align*}
where the notation $\expfam(x, \eta)$ denotes $x$ is drawn
from an exponential family with natural parameter $\eta$.\footnote{We are loose with the base measure $h$ as
it can be absorbed into the dominating measure.}

\begin{figure}[t]
  \begin{center}
 	\centering
    \includegraphics[width=.325 \textwidth]{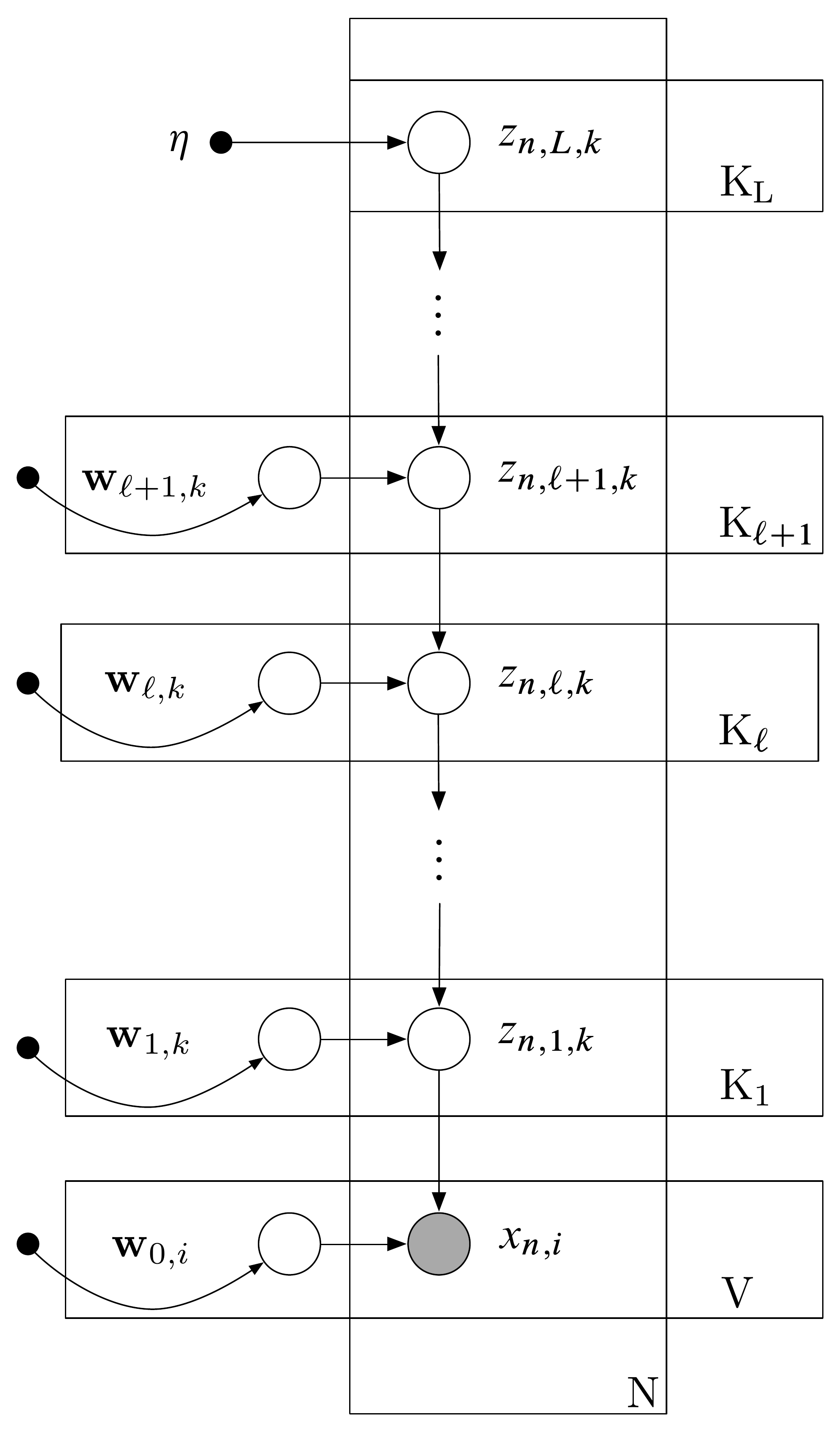}
     \caption{The deep exponential family with V observations.}
     \label{fig:gm}
 \end{center}
\end{figure}

Next, each latent variable
is drawn conditional on the previous layer,
\begin{equation}
  p(z_{\ell,k} \g \defbold{z}_{\ell +1},  \defbold{w}_{\ell,k}) =
  \expfam_\ell(z_{\ell, k}, g_\ell(\defbold{z}_{\ell+1}^\top \defbold{w}_{\ell,k})).
  \label{eq:exp_layer}
\end{equation}
The function $g_{\ell}$ maps the inner product to the natural
parameter.  Similar to the literature on generalized linear
models~\cite{Nelder:1972}, we call it the \textit{link function}. 
\myfig{gm} depicts this conditional structure in a graphical model.


Confirm that the
dimensions work: $z_{\ell, k}$ is a scalar; $\defbold{z_{\ell+1}}$ 
is a $K_{\ell+1}$ vector and  $\defbold{w_{\ell,
  k}}$ is a column from a $K_{\ell+1} \times K_{\ell}$ dimension matrix.
Note each of the $K_{\ell}$ variables
in layer $\ell$ depends on all the variables of the higher layer.
This gives the model the flavor of a neural network. The subscript $\ell$
on $\expfam$ indicates the type of exponential family can change across 
layers. The hierarchy of latent variables defines the DEF.  

DEFs can also be understood as random effects models~\cite{gelman2013bayesian} where the random variables are controlled by the product of a weight vector and a set of latent covariates.

\paragraph{Likelihood.}
The data are drawn conditioned on the lowest layer of the DEF, 
$p(x_{n,i} \g \defbold{z_{n,1}})$. Separating the likelihood from the
DEF will allow us to compose and embed DEFs in other models.
Later, we provide an example where we combine two DEFs to 
form a model for pairwise data.

In this paper we focus on count data, thus
we use the Poisson distribution as the observation likelihood. The Poisson
distribution with mean $\lambda$ is
\begin{align*}
p(x_{n, i} = x) = e^{-\lambda} \frac{\lambda^{x}}{x!}.
\end{align*}
If we let $x_{n, i}$ be the count of type $i$ associated with 
observation $n$, then $x_{n, i}$'s distribution is
\begin{align*}
 p(x_{n,i} \g \defbold{z}_{1}, \defbold{W}_{0}) = \textrm{Poisson}(\defbold{z}_{n, 1}^\top \defbold{w}_{0, i}),
\end{align*}
The observation weights $\defbold{W}_{0}$ is matrix where each entry is gamma distributed. We 
will discuss gamma distribution further in the next section.

Returning to the example from the introduction of modeling documents,
the $x_n$ are a vector of term counts. This means the observation weights
$\defbold{W}_0$ put positive mass on groups of terms. Thus, they form ``topics." 
Similarly, the weights on the second layer represents ``super topics," and 
the weights on the third layer represent ``concepts."
The distribution $p(\defbold{z}_{n, 1} \g \defbold{z}_{n,2}, \defbold{W}_1)$ represents the distribution of
``topics" given the ``super topics" of a document. \myfig{nyt_topics}
depicts the compositional and sharing semantics of DEFs.


\begin{figure}
  \begin{center}
 	\centering
    \includegraphics[width=.355 \textwidth]{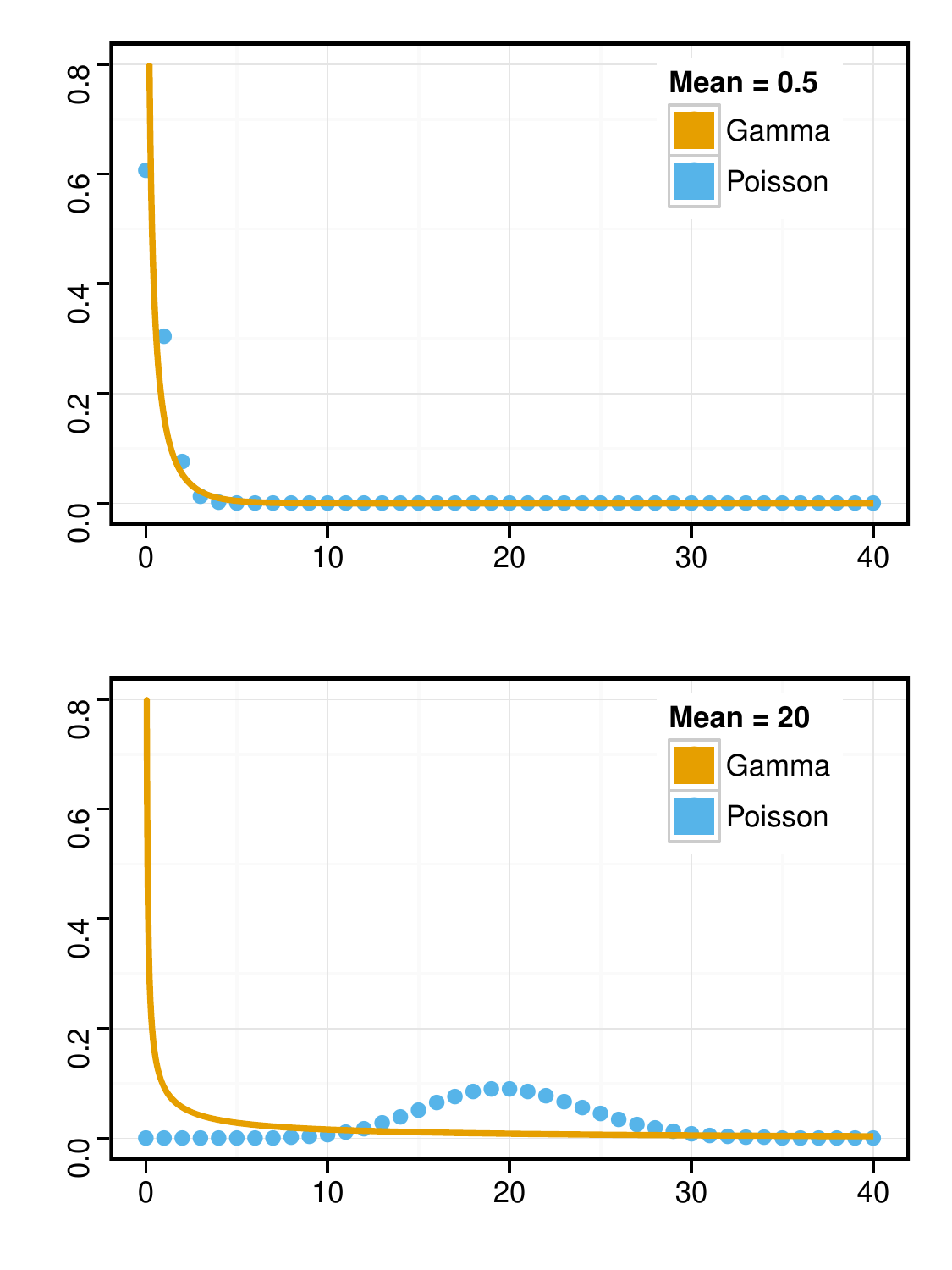}
     \caption{Draws from the Poisson (blue) and sparse gamma distribution (orange)
     				with low and high mean. The
     	           shape of the sparse gamma is held fixed. Note the high mean shifts the Poisson, while 
     	           does not shift the sparse gamma. Notice the spike-slab appearance
     	           of the sparse gamma distribution.}
     \label{fig:poisson_gamma}
 \end{center}
\end{figure}

\paragraph{The link function.}


Here we explore some of the connections between neural networks and deep 
exponential families.
As we discussed, the latent variable layers in deep exponential families are connected together via a
link function, $g_\ell$. Specifically the natural parameters for $z_{\ell, k}$ are 
specified by $g_\ell(\defbold{z}_{\ell+1}^\top \defbold{w}_{\ell, k})$.

Using properties of exponential families we can determine how the link
function alters the distribution of the $\ell$th layer. The moments
of the sufficient statistics of an exponential family are given by the
gradient of the log-normalizer $\nabla_\eta a(\eta)$. These moments
completely specify the exponential family \citep{Brown:1986}.
Thus in DEFs, the mean of the next layer is controlled by the link
function $g_\ell$ via the gradient of the log-normalizer,
\begin{align} 
\E[T(z_{\ell, k})] = \nabla_\eta a(g_l(\defbold{z}_{\ell+1}^\top \defbold{w}_{\ell,k})).
\label{eq:link_prop}
\end{align}
Consider the case of the identity link function, where $g_l(x) = x$. 
In this case, the expectation of latent variables in deep exponential families is transformed by
the log-normalizer at each level. This transformation of the expectation
is one source of non-linearity in DEFs. It parallels the non-linearities used in neural networks. 

To be clear, here is a representation of how the values and weights of one
layer control the expected sufficient statistics at the next:
\begin{minipage}{0.5\textwidth}
\begin{center}
    \includegraphics[width=0.5\columnwidth]{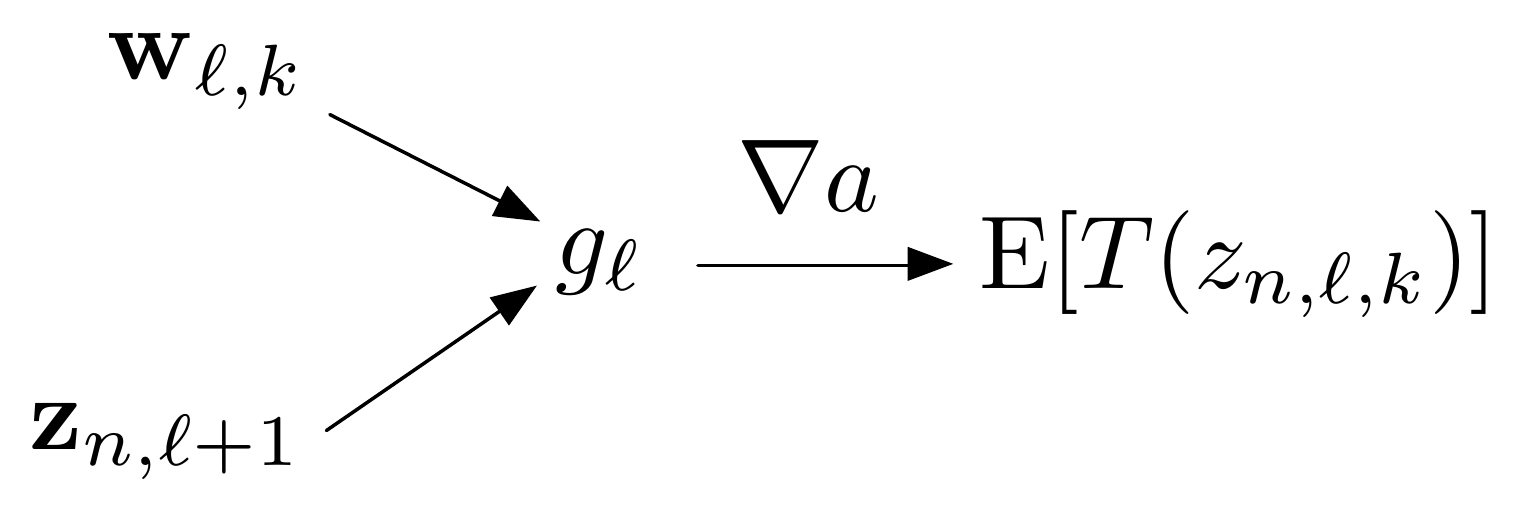}
\end{center}
\end{minipage}
\vspace{-0.1cm}

For example, in the sigmoid belief network \citep{Neal:1990},  we will see that the identity link
function recovers the sigmoid transformation used in neural networks.


\section{Examples}

To illustrate the potential of deep exponential families,
we present three examples: the sparse gamma DEF, the sigmoid belief
network, and a latent Poisson DEF. 


\paragraph{Sparse gamma DEF}
The sparse gamma DEF is a DEF with gamma distributed layers.
The gamma distribution is an exponential family distribution 
with support over the positive reals.
The probability density of the gamma distribution 
with natural parameters, $\alpha$ and $\beta$, is
\begin{align*}
p(z) = z^{-1}\exp(\alpha \log(z) - \beta z - \log \Gamma(\alpha) - \alpha  \log(\beta)).
\end{align*}
where $\Gamma$ is the gamma function. The expectation of the gamma distribution is
$\E[z] = \alpha \beta^{-1}$.

Through the link function in DEFs, the inner product of the previous layer and 
the weights control the natural parameters of the next layer. 
For sparse gamma models, we let components in a layer to control the expected
activation of the next layer, while the shape at each layer remains fixed.

Let $\alpha_\ell$ be the shape at layer $\ell$, then the link function for sparse gamma model is
\begin{align*}
g_\alpha = \alpha_\ell, \quad
g_\beta = \frac{\alpha_\ell}{\defbold{z}_{\ell+1}^\top \defbold{w}_{\ell,k}}.
\end{align*}
As the expectation of gamma variables needs to be positive, we let the weight matrices 
be gamma distributed as well.

When $\alpha$ is small (less than $1$), gamma distribution 
puts most of its mass near zero, and we call this type of distributions \emph{sparse gamma}. 
This type of distribution is akin to a soft spike-slab prior \citep{Ishwaran:2005}. 
Spike and slab priors have shown to perform well on feature selection
and unsupervised feature discovery \citep{Goodfellow:2012, Hernandez:2013}.
Thus sparse gamma distributions are like spike and slab priors, so we use them 
as the prior for the observation weights $W_1$ and all DEF weights that are 
constrained to be positive.

The sparse gamma distribution differs from distributions such as the normal and Poisson
in how the probability mass moves given a change in the mean.
For example, when the expected value is high, draws from the Poisson distribution
are likely to be much larger than zero, while in the sparse gamma distribution
draws will either be close to zero or very large.
This is 
like growing the slab of our soft spike and slab prior.
\myfig{poisson_gamma} visually 
demonstrates this. We plot both the Poisson and sparse gamma distribution in both settings. 
As an example in the sparse gamma DEF for documents, this means 
an observation does not have to express
every ``super topic" in a ``concept" it expresses.

We estimate the posterior on this DEF using one to three layers for two large text copora: \emph{Science} and
\emph{The New York Times} (NYT). We defer the discussion of the details of the corpora to \mysec{Experiments}. The topic hierarchy shown earlier in \myfig{nyt_topics} is from a three layer
sparse gamma DEF. 
In the appendix we present a portion of the \emph{Science} hierarchy.

\paragraph{Sigmoid belief network.}
The sigmoid belief network \citep{Neal:1990, Mnih:2014} is a widely used 
deep latent variable model. It consists of latent Bernoulli
layers where the mean of a feature at layer $\ell$ is given
by a linear combination of the features at layer $\ell+1$ with
the weights passed through the sigmoid function.

This is a special case of a deep exponential family with Bernoulli 
latent layers and the identity link function. To see this,
use \myeqp{exp_layer} to form the Bernoulli conditional 
of a hidden layer,
\begin{align*}
p&(z_{\ell,k} \g \defbold{z}_{\ell+1}, \defbold{w}_{\ell, k}) \\
&=  \exp(\defbold{z}_{\ell+1}^\top \defbold{w}_{\ell, k} z_{\ell, k} - \log(1 + \exp(\defbold{z}_{\ell+1}^\top \defbold{w}_{\ell,k})))
\end{align*}
where $z_{\ell,k} \in \{0,1\}$.

Using \myeqp{link_prop}, the expectation of $z_{\ell,k}$ is
the derivative of the log-normalizer of the Bernoulli. This derivative is the logistic function. Thus, 
a DEF with Bernoulli conditionals and identity link function recovers the sigmoid belief
network. The weights should be real valued, so, we set $p(\defbold{W}_\ell)$ to be a factorized normal distribution.

In the sigmoid belief network, we allow for the natural parameters to have intercepts. The intercepts provide a 
baseline activation for each feature independent of the shared weights.

%

\paragraph{Poisson DEF.}
The Poisson distribution is a distribution over counts that models
the number of events occurring in an interval. 
Its exponential family form with parameter $\eta$ is
\begin{align*}
p(z) = {z!}^{-1} \exp(\eta z - \exp(\eta)),
\end{align*}
where the mean of this distribution is $e^\eta$.

We define the Poisson deep exponential family 
as a DEF with latent Poisson levels and log-link function.~\footnote{We add an intercept to ensure
positivity of the rate.} 
This corresponds to the following conditional distribution
for the layers
\begin{align*}
p&(z_{\ell,k} \g \defbold{z}_{\ell+1}, \defbold{w}_{\ell, k}) \\
&= {(z_{\ell,k}!)}^{-1} \exp(\log(\defbold{z}_{\ell+1}^\top \defbold{w}_{\ell,k}) z_{\ell,k} - \defbold{z}_{\ell+1}^\top \defbold{w}_{\ell,k}).
\end{align*}
In the case of document modeling, the value of $z_{2,k}$ represents how many times ``super topic" k is
represented in this example.

Using the link function property of DEFs described earlier, the mean activation
at the next layer is given by the gradient of the log-normalizer
\begin{align*}
\E[z_{lk}] = \nabla_\eta a(\log(\defbold{z}_{\ell+1}^\top \defbold{w}_{\ell,k})).
\end{align*}
For the Poisson, $a$ is the exponential function. Thus its derivative is
the exponential function. This means the mean of the next layer is equal to
a linear combination of the weights of the previous layer. Our choice of link 
function requires positivity on the weights, so
we let $p(\defbold{W}_\ell)$ be a factorized gamma distribution.

We also consider Poisson DEFs with real valued weights to allow negative relations between lower layers and higher layers. 
We set the prior on the weights to be Gaussian in this case.
We use log-softmax $\eta=\log (\log (1+\exp(-\defbold{z}_{\ell+1}^T \defbold{w}_{\ell,k})))$ as the link function, where the function inside the first $\log$ is the softmax function. It preserves approximate linear relations between mean activation and the inner product $\defbold{z}_{\ell+1}^T \defbold{w}_{\ell,k}$ when it is large while allowing for the inner product to take negative values as well. 

Similar to the sigmoid belief network, we allow the natural parameter to have an intercept. This type of Poisson DEFs can be seen as an extension of the sigmoid belief network,
where each observation expresses an integer count number of a feature rather than just turning the feature
on or off. Table \ref{tab:dist-and-link} summarizes the DEFs we have described and will study in our experiments.

\begin{table*}
\centering
\begin{tabular}{|c | c |c | c | c | c |}
\hline
z-Dist & $\defbold{z}_{\ell+1}$ & W-dist & $\defbold{w}_{\ell, k}$ & $g_\ell$ & $\E[T(z_{\ell, k})]$ \\
\hline
Gamma & $R_+^{K_{\ell +1}} $ & Gamma & $R_+^{K_{\ell +1}}$ &  [constant; inverse] & $ [z_{\ell+1}^\top \defbold{w}_{\ell, k}$; $\Psi(\alpha_\ell) - \log (\alpha) + \log(z_{\ell+1}^\top \defbold{w}_{\ell, k})]$ \\
Bernoulli & $\{0, 1\}^{K_{\ell +1}}$ & Normal  & $R^{K_{\ell +1}}$ & identity & $\sigma(z_{\ell+1}^\top \defbold{w}_{\ell, k})$ \\
Poisson & $N^{K_{\ell +1}} $  & Gamma & $R_+^{K_{\ell +1}}$ &  log & $z_{\ell+1}^\top \defbold{w}_{\ell, k}$ \\
Poisson & $N^{K_{\ell +1}} $  & Normal & $R^{K_{\ell +1}}$ &  log-softmax & $\log( 1 + \exp(z_{\ell+1}^\top \defbold{w}_{\ell, k}))$ \\
\hline
\end{tabular}
\caption{
A summary of all the DEFs we present in terms of their layer distributions, weight distributions, and link functions.
}  \label{tab:dist-and-link}
\vskip -0.10in
\end{table*}

\section{Related Work}
Graphical models and neural nets have a long and
distinguished history. A full review is outside of the scope of this
article, however we highlight some key results as they relate to
DEFs. More generally, deep exponential families fall into the broad class of stochastic feed forward 
belief networks \citep{Neal:1990}, but \citet{Neal:1990} focuses mainly on one example in this class,
the sigmoid belief network, which is a binary latent variable model. Several existing stochastic feed 
forward networks are DEFs, such as latent Gaussian models \citep{Rezende:2014} and the sigmoid belief
network with layerwise dependencies \citep{Mnih:2014}. 

Undirected graphical models have also been used in inferring compositional
hierarchies. \citet{Salakhutdinov:2009b} propose
deep probabilistic models based on Restricted Boltzmann Machines (RBMs) \citep{Smolensky:1986}. 
RBMs are a two layer undirected probabilistic model with one layer of latent variables and one
layer of observations	 tied together by a weight matrix.
Directed models such as DEFs 
have the property of explaining away, where independent latent variables under the prior become
dependent conditioned on the observations. This property
makes inference harder than in RBMs, but forces a more parsimonious representation where
similar features compete to explain the data rather than work in tandem \citep{Goodfellow:2012, Bengio:2013}.

RBMs have been extended to
general exponential family conditionals in a model called exponential
family harmoniums (EFH) \citep{Welling:2004}. A certain infinite DEF with tied weights 
is equivalent to an EFH \citep{Hinton:2006}, but as our 
weights are not tied, deep exponential families
represent a broader class of models than exponential family harmoniums (and RBMs).

The literature of latent variable models relates to DEFs through hierarchical 
models and Bayesian factor analysis. Latent tree hierarchies have been
constructed with specific distributions (Dirchlet) \cite{Li:2006}, while
Bayesian factor analysis methods such as exponential family PCA \cite{Mohamed:2008} 
and multinomial PCA \citep{Buntine:2004} can be seen as a single layer deep exponential family.


\section{Inference}
The central computational problem for working with 
DEFs is posterior inference. The intractability 
of the partition function means posterior computations require
approximations. Past work on sigmoid belief networks has proposed doing greedy
layer-wise learning (for a specific kind of network) \cite{Hinton:2006}. 
Here, instead, we develop variational
methods \cite{Saul:1996} that are applicable to general DEFs.


Variational inference \citep{Jordan:1999} casts the posterior inference problem as 
an optimization problem. Variational algorithms seek to  minimize the KL divergence to the 
posterior from an approximating distribution $q$. This is equivalent to maximizing the following
\citep{Bishop:2006},
\begin{align*}
\cL (q)= \E_{q(z, W)}[\log p (x, z, W) - \log q (z, W)],
\end{align*}
where $z$ denotes all latent variables associated with each observation and 
$W$ all latent variables shared across observations.
This objective function is called the Evidence Lower BOund (ELBO) because it is a lower bound on
$\log p(x)$.

For the approximating distribution, $q$, we use the mean field variational family. 
In the mean field approximating family, the distribution over the latent variables factorizes.
We focus on the running example of a Poisson likelihood and $n$ observations.
The variational family is
\begin{align*}
q(z, W) = q(\defbold{W}_{0}) \prod_{\ell=1}^L q(\defbold{W}_{\ell}) \prod_{n=1}^N q(\defbold{z}_{n, \ell}),
\end{align*}
where $q(z_{n, \ell})$ and $q(W_{\ell})$  are fully factorized. Each component
in $q(\defbold{z}_{n, \ell})$ is
\begin{align*}
q(z_{n, \ell, k}) = \expfam_\ell(z_{n, \ell, k}, \lambda_{_{n,\ell, k}}),
\end{align*}
where the exponential family is the same one as the model distribution $p$. Similarly,
we choose $q(W)$ to be in the same family as $p(W)$ with parameters $\xi$. 

To maximize the ELBO, we need to compute expectations under the 
approximation $q$. These expectations for general DEFs will not have a simple
analytic form. Thus we use more recent ``black box" variational inference 
techniques that step around computing this expectation \citep{Wingate:2013, Salimans:2013, Ranganath:2014}.

Black box variational inference methods use stochastic optimization\citep{Robbins:1951}
to avoid the analytic intractability of computing the objective function. Stochastic optimization 
works by following noisy unbiased gradients. In black box variational inference \citep{Ranganath:2014}, the 
gradient of the ELBO with respect to the parameters of a latent variable can be written
as an expectation with respect to the variational approximation.

More formally, if we
let $p_{n, \ell, k} (x, z, W)$ be the terms in the log-joint that contains $z_{n, \ell, k}$ (its Markov
blanket), then the gradient for the variational approximation of $z_{n, \ell, k}$ is
\begin{multline*}
\nabla_{\lambda_{n,l,k}}\cL = E_{q}[\nabla_{\lambda_{n,\ell, k}}  \log q(z_{n, \ell, k}) 
\\(\log p_{n, \ell, k} (x, z, W) - \log q(z_{n, \ell, k}))].
\end{multline*}
We compute Monte Carlo estimates of this gradient by averaging the evaluation of
the gradient at several samples.
To compute the Monte Carlo estimate of the gradient, we need to be able to sample
from the approximation to evaluate the Markov 
blanket for each latent variable, the approximating distribution, and the gradient of the
log of the approximating distribution (score functions). We detail the score functions in the appendix.
From this equation, we can see that the primary cost in computing the gradients is in evaluating
the likelihood and score function on a sample. 
To speed up our algorithm, we parallelize the likelihood computation across
samples.

The Markov blanket for a latent variable in the first layer of a DEF is
\begin{align}
\log p_{n, 1, k} (x, z, W) &= \log p (z_{n, 1, k} | \defbold{z}_{n, 2},  \defbold{w}_{1, k}) \nonumber \\
&+ \log p (x_n | \defbold{z}_{n, 1}, \defbold{W}_{0}).
 \label{eq:mb1}
\end{align}
The Markov blank for a latent variable in the intermediate layer is
\begin{align}
\log p_{n, \ell, k} (x, z, W) =& \log p (z_{n, \ell, k} | \defbold{z}_{n, \ell + 1}, \defbold{w}_{\ell, k}) \nonumber \\
&+ \log p (\defbold{z}_{n, \ell - 1} | \defbold{z}_{n, \ell}, \defbold{W}_{\ell -1}).
 \label{eq:mb2}
\end{align}
The Markov blanket for the top layer is
\begin{align}
\log p_{n, L, k} (x, z, W) =& \log p (z_{n, L, k}) \nonumber \\
&+ \log p (\defbold{z}_{n, L - 1} | \defbold{z}_{n, L}, \defbold{W}_{L - 1}).
 \label{eq:mb3}
\end{align}
The gradients and Markov blankets for $W$ can be written similarly.

Stochastic optimization requires a learning rate to scale the noisy gradients before applying them to
the current parameters. We use RMSProp which scales the gradient by the square 
root of the online average of the sqaured gradient.\footnote{\tiny\url{www.cs.toronto.edu/~tijmen/csc321/slides/lecture_slides_lec6.pdf}}
 RMSProp captures the varying length scales and noise through 
the sum of squares term used to normalize the gradient step. We present a sketch of the algorithm in Algorithm 1,
and present the full algorithm in the appendix.

\begin{algorithm}[tb]
   \caption{BBVI for DEFs}
   \label{alg:def_small}
\begin{algorithmic}
 \STATE {\bfseries Input:} data $X$, model $p$, $L$ layers.
 \STATE {\bfseries Initialize} $\lambda, \xi$ randomly, $t = 1$.
  \REPEAT
 \STATE {Sample a datapoint $x$}
\FOR{s = 1 to $S$} 
\STATE{$z_x[s], W[s] \sim q$}
\STATE{$p[s] = \log p(z_x[s], W[s], x)$ }
\STATE{$q[s] = \log q(z_x[s], W[s])$ }
\STATE{$g[s] = \nabla \log q(z_x[s], W[s])$ }
\ENDFOR
\STATE{Compute gradient using BBVI}
\STATE{Update variational parameters for $z$ and $W$}
 \UNTIL{change in validation likelihood is small }
\end{algorithmic}
\vskip -0.05in
\end{algorithm}

\section{Experiments}
\label{sec:Experiments}

We have introduced DEFs and detailed a procedure for posterior
inference in DEFs. We now provide an extensive evaluation of DEFs. 
We report predictive results from 28 different DEF instances where we explore
the number of layers (1, 2 or 3), the latent variable distributions (gamma,
Poisson, Bernoulli) and the weight distributions (normal, gamma) using a
Poisson observational model. Furthermore, we instantiate and report results
using a combination of two DEFs for pairwise data. 

Our results: 
\begin{itemize}[noitemsep,topsep=0pt,parsep=0pt,partopsep=0pt]
\item Show improvements over strong baselines for both topic modeling and
collaborative filtering on a total of four corpora. 
\item Lead us to conclude that deeper DEFs and sparse gamma DEFS display the
strongest performance overall.
\end{itemize}

\subsection{Text Modeling}

We consider two large text corpora  \emph{Science} and
\emph{The New York Times}. \emph{Science} consists of 133K
documents and 5.9K terms. \emph{The New York Times} consists
of 166K documents and 8K terms.

\paragraph{Baselines.}
As a baseline we consider Latent Dirichlet Allocation \citep{Blei:2003} a
popular topic model, and state-of-the-art DocNADE \citep{Larochelle:2012}.
DocNADE estimates the probability of a given word in a document given the
previously observed words in that document. In DocNADE, the connections between
each observation and the latent variables used to generate the observations are
shared.

We note that the one layer sparse gamma DEF is equivalent to Poisson matrix
factorization~\citep{Canny:2004,Gopalan:2013} but our model is fully Bayesian
and our variational distribution is collapsed. 


\begin{table}
\centering
\begin{tabular}{|c | c " c | c|}
\hline
 Model & DEF \defbold{W} & \emph{NYT} & \emph{Science} \\
 \hline
 LDA~\citep{Blei:2003b}& & 2717 & 1711 \\
 DocNADE~\citep{Larochelle:2012} && 2496 & 1725 \\\hline
 Sparse Gamma 100  &$\emptyset$& 2525 & 1652 \\
 Sparse Gamma 100-30 & $\Gamma$ & 2303 & 1539 \\
 Sparse Gamma 100-30-15 &$\Gamma$& 2251 & 1542 \\\hline
 Sigmoid 100 &$\emptyset$& 2343 & 1633 \\
 Sigmoid 100-30 &$\mathcal{N}$&  2653    & 1665 \\
 Sigmoid 100-30-15 &$\mathcal{N}$& 2507 & 1653 \\\hline
 Poisson 100 &$\emptyset$& 2590 & 1620 \\
 Poisson 100-30 &$\mathcal{N}$& 2423 & 1560 \\
 Poisson 100-30-15 &$\mathcal{N}$& 2416 & 1576 \\
 Poisson  log-link 100-30 &$\Gamma$& 2288 & 1523 \\
 Poisson  log-link 100-30-15 &$\Gamma$& 2366 & 1545 \\
\hline
\end{tabular}
\caption{Perplexity on held out collection of 1K \emph{Science} and \emph{NYT}
documents. Lower values are better. The DEF $\defbold{W}$ column indicates the
type of prior distribution over the DEF weights, $\Gamma$ for the gamma prior
and $\mathcal{N}$ for normal (recall that one layer DEFs consist only of a
layer of latent variables, thus we represent their prior with the $\emptyset$).  
} 
\label{tab:text_results}
\vskip -0.10in
\end{table}

\begin{table*}
\centering
\begin{tabular}{|c | c | c | c | c |}
\hline
Model & Netflix Perplexity & Netflix NDCG & ArXiv Perplexity & ArXiv NDCG  \\
\hline
 Gaussian MF~\cite{Salakhutdinov:2008a}  & -- & 0.008 & --  & 0.013 \\\hline
 1 layer Double DEF & 2319  & 0.031 & 2138  & 0.049 \\ 
 2 layer Double DEF & 2299 & 0.022 & 1893  & 0.050 \\
 3 layer Double DEF & 2296& 0.037  & 1940  & 0.053 \\
\hline
\end{tabular}
\caption{A comparison of a matrix factorization methods on Netflix and the
ArXiv. We find that the Double DEFs outperform the shallow ones on 
perplexity. We also find that 
the NDCG of around 100 low-activity users (users with less than
5 and 10 observations in the observed 10\% of the held-out set respectively for
Netflix and ArXiv). We use Vowpal Wabbit's MF implementation which does not
readily provide held-out likelihoods and thus we do not report the perplexity
associated with MF.}  
\label{tab:movie_results}
\vskip -0.10in
\end{table*}

\paragraph{Evaluation.}
We compute perplexity on a held out set of 1,000
documents. Held out perplexity is given by
\begin{align*}
\exp{\left(\frac{-\sum_{d \in \textrm{docs}} \sum_{w \in d} \log p(w \g \text{\# held out in } d )}{N_\textrm{held out words}} \right)} 
\end{align*}
Conditional on the total number of held out words, the distribution of the held out words becomes
multinomial. The mean of the conditional multinomial is given by the normalized Poisson
rate in each document. We set the rates to the expected value under the variational distribution.
Additionally, we let all methods see ten percent of the words in each document; the other
ninety percent form the held out set. This similar to the document completion evaluation
metric in \citet{Wallach:2009a} except we query the test words independently.
We use the observed ten percent to compute the variational distribution 
for the document specific latent variables, the DEF for the document, while keeping the approximation on
the shared weights fixed. In DocNADE, this corresponds
to always seeing a fixed set of words first, then evaluating each new word given the first ten 
percent of the document. 

Held out perplexity differs from perplexity computed from the predictive distribution
$p(x^* \g x)$. The latter can be a more difficult problem as we only ever condition on a fraction
of the document. Additionally computing perplexity
from the predictive distribution requires computationally demanding sampling procedures which for 
most
models like LDA only allow testing of only a small number (50) of documents \citep{Wallach:2009a, Salakhutdinov:2009a}. 
In contrast our held-out test metric can be quickly computed for 1,000
test documents.

\paragraph{Architectures and hyperparameters.}
We build one, two and three layer hierarchies of the sparse gamma DEF,
the sigmoid belief network, Poisson DEF, and log-link Poisson DEF. The sizes
of the layers are 100, 30, and 15, respectively. We note that while
different DEFs may have better predictive performance at different sizes, 
we consider DEFs of a fixed size as we also seek a compact explorable representation
of our corpus. One hundred topics fall into the range of topics searched in the topic modeling
literature \cite{Blei:2007}. We detail the hyperparameters for each DEF in the appendix.

We observe two phases to DEF convergence, it converges quickly to a good held-out
perplexity (around 2,000 iterations) and then slowly improves until final
convergence (around 10,000 iterations). Each iteration takes approximately 30
seconds on a modern 32-core machine (from Amazon's AWS). 

\paragraph{Results.}
Table \ref{tab:text_results} summarizes the predictive results on both corpora.
We note that DEFs outperform the baselines on both datasets. 
Furthermore moving
beyond one layer models generally improves performance as expected.  
The table also reveals that stacking layers of gamma latent variables leads always leads to
similar or better performance.
Finally, as shown by the Poisson DEFs with different link functions, we find
gamma-distributed weights to outperform normally-distributed
weights. Somewhat related, we find sigmoid DEFs (with normal weights) to be more difficult
to train and deeper version perform poorly. 

%

\subsection{Matrix Factorization}
Previously, we constructed models out of a single DEF, but DEFs can be
embedded and composed in more complex models. We now present the double
DEF, a factorization model for pairwise data where both the rows and
columns are determined by DEFs.	The graphical model of this double DEF
corresponds to replacing $\defbold{W}_{0}$ in \myfig{gm} with another DEF.

We focus on factorization of counts (ratings, clicks). The observed data
are generated with a Poisson distribution.
The observation likelihood for this double DEF is $p(x_{n,i} \g \defbold{z}_{n, 1}^c, \defbold{z}_{i, 1}^r) =
 \textrm{Poisson}({\defbold{z}_{n, 1}^c}^\top {\defbold{z}_{i, 1}^r})$,
where $\defbold{z}_{n, 1}^c$ is the lowest layer of a DEF for the $n$th
observation and $\defbold{z}_{i, 1}^r$ is the lowest layer of a DEF for
the $i$th item. 
The double DEF has hierarchies on both users and items.

We infer double DEFs on \emph{Netflix} movie ratings and click data from the
\emph{ArXiv} ({\small\url{www.arXiv.org}}) which indicates
how many times a user clicked on a paper.  Our \emph{Netflix} collection
consists of 50K users and 17.7K movies.
The movie ratings range from zero to five stars, where zero means the movie was unrated by the user.
The \emph{ArXiv} collection consists of 18K users and 20K documents.
We fit a one, two, and three layer double DEF where the sizes of the row DEF match the 
sizes of the column DEF at each layer. The sizes of the layers are 100,
30, and 15. We compare double DEFs to $l2$-regularized (Gaussian) matrix
factorization (MF)~\citep{Salakhutdinov:2008a}.  We re-use the testing procedure
introduced in the previous section (this is referred to as
\textit{strong-generalization} in the recommendation
literature~\citep{Marlin:04}) where the held-out test set contains one thousand
users. For performance and computational reasons we subsample
zero-observations for MF as is standard~\cite{Gopalan:2013}.  
Further, we also report the commonly-used multi-level ranking
measure (un-truncated) NDCG \citep{Jarvelin:2000} for all methods. 

Table \ref{tab:movie_results} shows that two-layer DEFs improve performance
over the shallow DEF and that all DEFs outperform Gaussian MF. On perplexity the three layer
model performs similarly on Netflix and slightly worse on the ArXiv.
The table further highlights that when comparing ranking performance, the
advantage of deeper models is especially clear on low-activity users (NDCG across all
test users is comparable within the three DEFs architectures and is not
reported here). This data regime is of particular importance for practical
recommender systems.  We postulate that this due to the hierarchy in deeper
models acting as a more structured prior compared to single-layer models.





\section{Discussion}
We develop deep exponential families as a way to describe hierarchical relationships of
latent variables to capture compositional semantics of data. 
We present several instantiations of deep exponential families and achieve
improved predictive power and interpretable semantic structures for both problems in
text modeling and collaborative filtering.

{\small
\bibliography{../bib.bib}
}

\section*{Appendix}

\paragraph{General Algorithm.}
Following the notation from the main paper, the general algorithm
for mean field variational inference in deep approximating families
is given in (Alg. \ref{alg:full_def}).

\begin{algorithm}[tb]
   \caption{BBVI for DEFs}
   \label{alg:full_def}
\begin{algorithmic}
 \STATE {\bfseries Input:} data $X$, model $p$, $L$ layers .
 \STATE {\bfseries Initialize} $\lambda, \xi$ randomly, $t = 1$.
  \REPEAT
 \STATE {\bfseries // Draw a single data point from X }
 \STATE $n = $Unif$(D)$
\STATE {\bfseries // Get S samples in parallel }
\FOR {$s=1$ {\bfseries to} S}
 \STATE $z_1[s] \sim q(z_1 ; \lambda_{n, 1})$
 \STATE $W_0[s] \sim q(W_0 \g \xi_0)$
 \STATE $p_0[s] = \log p(x_n \g z_1[s] , W_0[s])$
 \STATE $q_1[s] = \log q(z_1[s] ; \lambda_{n, 1} )$
 \STATE $g_1[s] =\nabla_{\lambda_{n, 1}} \log q(z_1[s] ; \lambda_{n, 1} )$ 
\STATE $g_{W_0}[s] = \nabla_{\xi_{1}} \log q(W_{0})$
 \STATE $p_{W_0}[s] = \log p(W_0; \xi_1)$
 \STATE $q_{W_0}[s] = \log q(W_0; \xi_1)$
  \FOR{$l=2$ {\bfseries to} L}
  	\STATE $z_l[s] \sim q(z_l ; \lambda_{n, l} )$
  	 \STATE $W_{l-1}[s] \sim q(W_{l-1} \g \xi_{l-1})$
  	\STATE $p_{l}[s] = \log p(z_{l-1} \g z_l, W_{l-1}[s])$
  	\STATE $q_l[s] =\log q(z_l ; \lambda_{n, l} )$
	\STATE $g_l[s] = \nabla_{\lambda_{n, l}} \log q(z_l; \lambda_{n, l} )$
         \STATE $g_{W_{l-1}}[s] = \nabla_{\xi_{l-1}} \log q(W_{l-1})$
          \STATE $p_{W_{l-1}}[s] = \log p(W_{l-1}; \xi_{l-1})$
          \STATE $q_{W_{l-1}}[s] = \log q(W_{l-1}; \xi_{l-1})$
\ENDFOR
\STATE $p_{L}[s] = \log p (z_L)$
\ENDFOR
\STATE {\bfseries // Update parameters} 
 \FOR{$l=1$ {\bfseries to} L}
   \FOR {$k=1 $ {\bfseries to} $K_l$}
    \STATE $S = g_{l, k} (p_{l-1} + p_{l, k} - q_{l, k})$
    \STATE $\lambda_{n, 1, k} = \lambda_{n, 1, k} + \rho$ mean$(S)$
 \ENDFOR
\STATE $T = g_{W_{l-1}} (p_{W_{l-1}} - q_{W_{l-l}} + p_{l-1})$
\STATE $\xi_{l-1} = \xi_{l - 1}  + \rho$ mean$(T)$
 \ENDFOR
 \UNTIL{change of val likelihood is less than 0.01.}
\end{algorithmic}
\vskip -0.05in
\end{algorithm}

\paragraph{Properties of $q$.}
In our experiments, we use four variational families (Poisson, gamma,
Bernoulli, and normal). We detail the necessary score functions here.
For the Poisson, the distribution is given by:
\begin{align*}
q(z) = e^{-\lambda} \frac {\lambda^z} {z!}.
\end{align*}

The score function is
\begin{align*}
\frac{\partial \log q(z)}{\partial \lambda} = -1 + \frac{z}{\lambda}.
\end{align*}

For the gamma, we use the shape $\alpha$ and scale $\theta$ as variational
parameters. The distribution is given by
\begin{align*}
q(z) = \frac{1} {\Gamma(\alpha)\theta^{\alpha}} z^{\alpha-1} e^{-z/\theta}.
\end{align*}
The score function is
\begin{align*}
\frac {\partial \log q(z)} {\partial \alpha} &= -\Psi(\alpha) - \log \theta,
+ \log z \\
\frac {\partial \log q(z)} {\partial \theta} &= - \alpha / \theta + z / \theta^2,
\end{align*}
where $\Psi$ is the digamma function.

For the Bernoulli distribution, we use the natural parameterization with parameter
$\eta$ to form the variational approximation. The distribution is
\begin{align*}
q(z) = \frac 1 {1+e^{-(2z-1)\eta}}.
\end{align*}
The score function is
\begin{align*}
\frac {\partial \log q(z)} {\partial \eta} = (2z-1)\frac
{e^{-(2z-1)\eta}} {1+e^{-(2z-1)\eta}}.
\end{align*}

For the normal variational approximation, we use the standard parameterization by
mean $\mu$ and variance $\sigma^2$. The distribution is
\begin{align*}
q(z) = \frac 1 {\sqrt{2\pi \sigma^2}} \exp{(-\frac{(x-\mu)^2} {2\sigma^2})}.
\end{align*}
The score function is
\begin{align*}
\frac {\partial \log q(z)} {\partial \mu} &= \frac {x-\mu} {\sigma^2} \\
\frac {\partial \log q(z)} {\partial \sigma^2} &= - \frac 1
{2\sigma^2} + \frac {(x-\mu)^2} {2\sigma^4}.
\end{align*}

\paragraph{Parameterizations of Variational Distributions.} 
Several of our variational parameters like the variance of the normal
have positive constraints To enforce positivity constraints, we transform
an unconstrained variable by $\log(1+\exp(x))$. To avoid numerical issues
when sampling, we truncate values when appropriate.
Gradients of the unconstrained parameters are obtained 
with the chain rule from the score
function and the derivative of softmax: $\exp(x)/(1+\exp(x))$.

\paragraph{Optimization} We perform gradient ascent step on the ELBO
using
\begin{equation}
\Delta \theta =  \rho \Gamma \nabla_{\theta} \text{ELBO}
\end{equation}
$\rho$ is a fixed scalar set to $0.2$ in our
experiments. $\nabla_{\theta} \text{ELBO}$ is a noisy gradient
estimated using BBVI. $\Gamma$ is a diagonal preconditioning matrix
estimated using the RMSProp heuristic\footnote{Described by G. Hinton,
  RMSprop: Divide the gradient by a running average of its recent
  magnitude, in Coursera online course: Neural networks for machine
  learning, lecture 6e, 2012.}. A diagonal element of $\Gamma$ is the
reciprocal of the squared root of a running average of the squares of
historical gradients of that component. We used a window size of $10$
in our experiments.

\begin{figure*}
	\centering
	    \includegraphics[width=\textwidth]{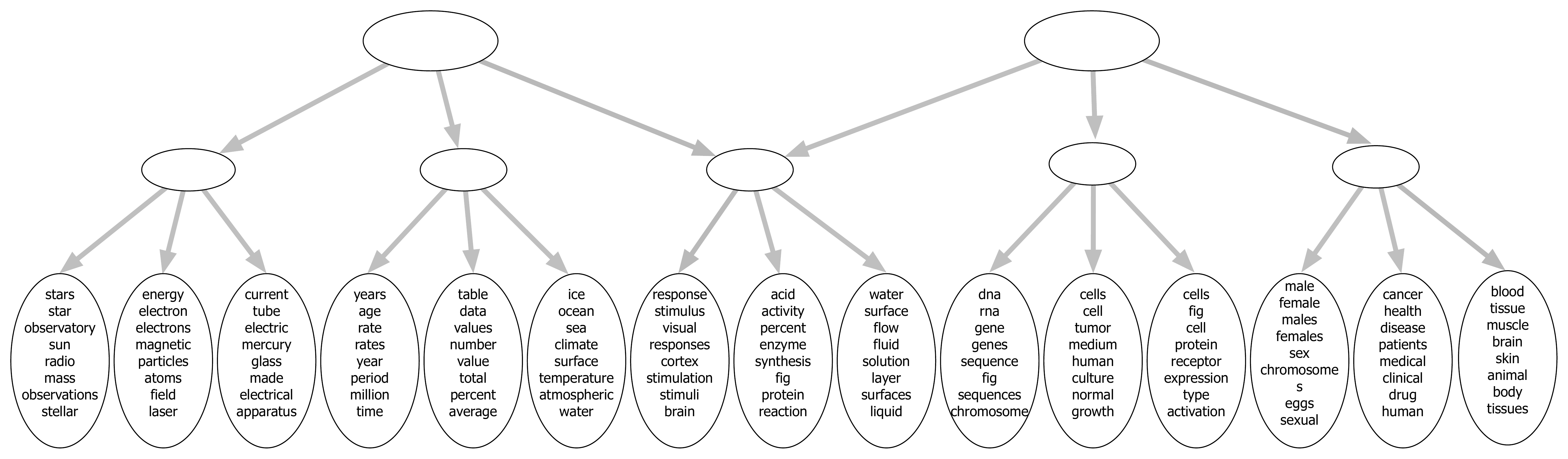}
	\caption{A fraction of the three layer topic hierarchy of the \emph{Science} corpus. The top words are shown for each ``topic."
	   			The arrows represent hierarchical groupings. We choose top
                three components at each layer. Similar ``topics" are grouped into ``super topics."  The two ``concepts" share a ``super topic."}
\end{figure*}

\paragraph{Hyperparameters and Convergence}
We use the same hyperparameters on Gamma distributions on each layer with
shape and rate $0.3$.
For the sigmoid belief network we use a prior of $0.1$ to achieve some sparsity as
well. We fix the Poisson prior rate to be $0.1$.
For gamma $W$'s we use shape $0.1$ and rate $0.3$.
For Gaussian $W$'s we use a prior mean of $0$ and variance of $1$.
We let the experiments run for 10,000 iterations at which point the
validation likelihood is stable.

For the double DEF, we set all shapes to $0.1$ and rates to $0.3$.  We let
the Double DEF experiment run for about 10,000 iterations. The validation
likelihood had converged for all models by this point.

\end{document}

%% file: preamble.tex
\usepackage{mdwlist} 
\usepackage{amsmath} 
\usepackage{epsfig}
\usepackage{color}
\usepackage{tablefootnote}
\usepackage{wrapfig}

\newcommand{\mysec}[1]{Section~\ref{sec:#1}}

\newcommand{\myeqp}[1]{Eq.~\ref{eq:#1}}

\newcommand{\myfig}[1]{Figure~\ref{fig:#1}}

\newcommand{\g}{\,\vert\,}
\newcommand{\E}{\textrm{E}}

\newcommand{\cL}{{\cal L}}

\newcommand{\expfam}{\textsc{expfam}}
\newcommand{\defbold}[1]{\ensuremath{\boldsymbol{\mathbf{#1}}}}